\documentclass[sigconf]{acmart}

\usepackage{booktabs} % For formal tables
\usepackage{multirow}
\newcommand\Tstrut{\rule{0pt}{2.6ex}}         % = `top' strut
\newcommand\Bstrut{\rule[-0.9ex]{0pt}{0pt}}   % = `bottom' strut

\usepackage{makecell}
\usepackage{amsfonts}
\usepackage{amsmath}
\usepackage{pgfplots} % for plot bar chart
\newcommand{\floor}[1]{\lfloor #1 \rfloor}

\usepackage{bm}

\usepackage{listings}
\usepackage{color}
\definecolor{codegreen}{rgb}{0,0.6,0}
\definecolor{codegray}{rgb}{0.5,0.5,0.5}
\definecolor{codepurple}{rgb}{0.58,0,0.82}
\definecolor{backcolour}{rgb}{0.95,0.95,0.92}

\usepackage{tikz}
\usetikzlibrary{matrix}
\usetikzlibrary{patterns}
\usetikzlibrary{shapes.geometric, arrows}
\tikzstyle{process} = [rectangle, minimum width=1cm, minimum height=1cm, text centered, draw=black ]
\tikzstyle{io} = [rectangle, minimum width=0.5cm, minimum height=0.1cm, text centered, draw=white]
\tikzstyle{up} = [yshift=0.2cm]
\tikzstyle{arrow} = [thick,->]
\tikzstyle{sq} = [rectangle, minimum width=0.7cm, minimum height=0.7cm, text centered, draw=black ]
\tikzstyle{cir} = [circle, minimum width=0.7cm, minimum height=0.7cm, text centered, draw=black ]
\tikzstyle{sqvec} = [matrix,draw=black,thick]
\tikzstyle{cirsmall} = [circle, minimum width=0.3cm, minimum height=0.3cm, draw=black ]
\tikzstyle{sqvec2} = [matrix,draw=black,thick]
\tikzset{
	sq1/.style={rectangle, minimum width=1cm, minimum height=1cm, text centered, draw=black},
	sq1m/.style={rectangle, minimum width=0.8cm, minimum height=0.4cm, text centered, draw=black},
	sq1p/.style={rectangle, minimum width=1cm, minimum height=1cm, text centered, draw=black, pattern=north west lines},
	circ/.style={circle, minimum width=0.4cm, minimum height=0.4cm, text centered, draw=black},
	arrow/.style={thick,->},
	sqvec/.style={matrix,matrix of nodes,nodes in empty cells},
}

\lstdefinestyle{mystyle}{
	backgroundcolor=\color{backcolour},
	commentstyle=\color{codegreen},
	keywordstyle=\color{magenta},
	numberstyle=\tiny\color{codegray},
	stringstyle=\color{codepurple},
	basicstyle=\footnotesize,
	breakatwhitespace=false,
	breaklines=true,
	captionpos=b,
	keepspaces=true,
	numbers=left,
	numbersep=5pt,
	showspaces=false,
	showstringspaces=false,
	showtabs=false,
	tabsize=2
}
\setcopyright{acmcopyright}
\acmDOI{10.475/123_4}

\acmISBN{123-4567-24-567/08/06}

\acmConference[SoICT '17]{Eighth International Symposium on Information and Communication Technology}{December 7--8, 2017}{Nha Trang City, Viet Nam}
\copyrightyear{2017}
\acmYear{2017}

\acmPrice{15.00}

\copyrightyear{2017}
\acmYear{2017}
\setcopyright{acmcopyright}
\acmConference[SoICT '17]{Eighth International Symposium on Information and Communication Technology}{December 7--8, 2017}{Nha Trang City, Viet Nam}
\acmPrice{15.00}
\acmDOI{10.1145/3155133.3155158}
\acmISBN{978-1-4503-5328-1/17/12}

\begin{document}
\title[Combining CNN and RNN for Sentiment~Analysis]{Combining Convolution and Recursive~Neural~Networks for Sentiment~Analysis} % long title and short title

\author{Vinh D. Van}
\orcid{1234-5678-9012}
\affiliation{%
	\institution{Faculty of Information Technology
		VNUHCM - University of Sciences }
	\streetaddress{Nguyen Van Cu}
	\city{Ho Chi Minh}
	\country{Vietnam}}
\email{vdvinh@apcs.vn}

\author{Thien Thai}
\affiliation{%
	\institution{Faculty of Information Technology
		VNUHCM - University of Sciences }
	\streetaddress{Nguyen Van Cu}
	\city{Ho Chi Minh}
	\country{Vietnam}}
\email{tthien@apcs.vn}

\author{Minh-Quoc Nghiem}
\affiliation{%
 \institution{Faculty of Information Technology
 VNUHCM - University of Sciences }
 \streetaddress{Nguyen Van Cu}
 \city{Ho Chi Minh}
 \country{Vietnam}}
\email{nqminh@fit.hcmus.edu.vn}

\renewcommand{\shortauthors}{V. Van et al.}

\begin{abstract}
This paper addresses the problem of sentence-level sentiment analysis.
In recent years, Convolution and Recursive Neural Networks have been proven to be effective network architecture for sentence-level sentiment analysis.
Nevertheless, each of them has their own potential drawbacks.
For alleviating their weaknesses, we combined Convolution and Recursive Neural Networks into a new network architecture.
In addition, we employed transfer learning from a large document-level labeled sentiment dataset to improve the word embedding in our models.
The resulting models outperform all recent Convolution and Recursive Neural Networks.
Beyond that, our models achieve comparable performance with state-of-the-art systems on Stanford Sentiment Treebank.
\end{abstract}




\maketitle

\section{Introduction}
In recent years, thanks to the dramatic growth of social media, customers' opinions are expressed in the highest speed and volume ever recorded in history.
It is inefficient to read, analyze, or even collect such a large amount of data manually.
Sentiment analysis offers a way to collect and process public opinion automatically.
Basically, sentiment analysis is used to determine whether an opinion about a specific product, event, or organization is positive or negative.
Formally, given document $d$, the main objective of sentiment analysis is to extract the following quintuple~\cite{liu2012sentiment}:
\[ ( e_{i}, a_{ij}, s_{ijkl}, h_{k}, t_{l} ) \]
Where:
\begin{itemize}
	\item $e_{i}$: entity \(i\) (entity extraction and categorization)
	\item $a_{ij}$: aspect \(j\) of entity \(i\) (entity extraction and categorization)
	\item $h_{k}$: holder \(k\) (opinion holder extraction and categorization)
	\item $t_{l}$: time \(l\) (time extraction and standardization)
	\item $s_{ijkl}$: opinion of holder \(k\) about aspect \(j\) of entity \(i\) at time \(l\) (aspect sentiment classification)
\end{itemize}
Sentence-level sentiment analysis is to determine whether a sentence expresses positive or negative sentiment.
This level of analysis assumes that every sentence contains one opinion toward an entity (e.g., a single movie)~\cite{liu2012sentiment}.

In this paper, we explore two ideas: ``Combining Convolution and Recursive Neural Networks'' (the main idea) and ``Transfer Learning From Large Review Dataset'' (the supporting idea).
\paragraph{Combining Convolution and Recursive Neural Networks} Convolution (CNN) and Recursive Neural Networks (RecNN) have been proven to be effective network architectures for sentence-level sentiment analysis.
Nevertheless, each of them has drawbacks (Section~\ref{sec:related}).
For alleviating their weaknesses, we combined CNN and RecNN into a new network architecture (Section~\ref{sec:cnn-treelstm}) which is able to outperform both CNN and RecNN (Section~\ref{sec:result}) on Stanford Sentiment Treebank (Section~\ref{sec:sst}).
This approach is closely related to the paper of Wang et al.~\cite{cnn-rnn} which investigates the of combination CNN and Recurrent Neural Networks (RNN).
\paragraph{Transfer Learning From Large Review Dataset} One obstacle of solving sentence-level sentiment analysis is the lack of labeled data which potentially causes many drawbacks, one of which is over-fitting word embedding (Section~\ref{sec:we-motive}).
Since most opinions are expressed in form the of  a documents (i.e., multiple sentences), sentence-level labeled dataset required more work to produce.
Until now, the largest dataset for sentence-level sentiment analysis is Standford Sentiment Treebank which only contains 11,855 sentences.
The number is insignificant compared to Amazon Review dataset which has 83.68 million reviews.
We utilized Amazon Review dataset to train a new word embedding (Section~\ref{sec:glove-amazon}) named ``Glove Amazon''.
By replacing the standard Glove\footnote{Common Crawl (840B tokens, 2.2M vocab, cased, 300d vectors, 2.03 GB download) publicly available at \url{https://nlp.stanford.edu/projects/glove/}} with Glove Amazon, many models can gain considerable improvements when evaluated on Stanford Sentiment Treebank.
We also demonstrated that a combination of Glove Amazon and standard Glove is better than each of the word embeddings (Section~\ref{sec:result}).

We provide the source code for the model as well as trained word vectors at \url{https://github.com/ttpro1995/Tree_CNN_LSTM}.

\section{Related Work and Motivation}\label{sec:related}
In recent years, sentiment analysis has enjoy dramatic improvements by applying different variations of CNN, RNN and RecNN.

Although originally invented for Computer Vision, CNNs have been proven to be effective models for document classification.
Nevertheless, for composing fixed-length representation vector of documents, several types of max pooling layers were employed~\cite{nlp-scratch, KimCNN, DCNN, 2-layer-cnn}.
Although max pooling layer largely simplified the network (which is good for preventing over-fit), this solution have a clear disadvantage.
By down-sampling a feature map, although the resulted vector still contains information about the existence of a feature, it is likely that the information about the position or order of the feature is lost.
This can be harmful because the order of words and phrases is important for understanding the sentiments of a sentence.

RNN and RecNN have been especially designed for dealing with variable-length sequential input.
They have been successfully applied to a variety of NLP tasks, include: speech recognition~\cite{speech-lstm, MiaoGM15}, sentiment analysis~\cite{treeLSTM, attention-gru}, text summarization~\cite{RushCW15, NallapatiXZ16}, machine translation~\cite{FiratCB16, SutskeverVL14, BritzGLL17}, language modeling~\cite{mikolov-nlm, JozefowiczVSSW16}.

Wang et al.~\cite{cnn-rnn} successfully combined CNN with RNN for Sentiment Analysis by utilizing CNN for capturing phrase-level features and RNN for composing these features.
Their network architecture was similar to that of Yoon Kim et al.~\cite{KimCNN}, the main different was the usage of RNN for replacing the max-over-time pooling layer.
Nevertheless, it was able to significantly outperform the state-of-the-art system at the time~\cite{cnn-rnn}.

In this paper, we adopting the idea of Wang et al.~\cite{cnn-rnn} by combining CNN with RecNN.
The main different is that the RNN module is replaced by a RecNN module.
RecNN have several advantages over RNN:
\begin{itemize}
	\item In case the input sequence belongs to a recursively defined language, given only a small subset of the data with limited length sentences, tree structures model have better ability to generalize comparing to sequential ones.
	However, when the limited length of sentences in the training data is increased, the advantage of tree over sequential models decreases fast~\cite{bowman-treevslstm}.
	\item Tree can break down complicated sentences into simpler phrases which are easier for generalization~\cite{knowledge-matter}~\cite{need-tree}.
	\item Some features which are far apart when a sentence is presented as sequence become closer when it is presented as a tree~\cite{need-tree}.
\end{itemize}

For our RecNN module, Tree-LSTMs~\cite{treeLSTM} was employed.
The core idea behind the design of Tree-LSTMs are to generalize the LSTM for tree-structured inputs.
Tree-LSTMs were able to achieve state-of-the-art performance on two tasks: predicting the semantic relatedness of two sentences (SemEval 2014, Task 1~\cite{SemeEvalTask1}) and sentiment classification (Stanford Sentiment Treebank~\cite{socher2013recursive}).
Nevertheless, Tree-LSTMs also have some drawbacks, including:
\begin{itemize}
	\item Sentences can be wrongly parsed, especially when comments are expressed in informal language.
	The performance of the system depends on the parser being used.
	\item At their leaf-module, Tree-LSTMs have only a simple logistic regression layer on top of the vector presentation of a single word at that position.
	The simple leaf-module of Tree-LSTMs might be its weakness when dealing with the problem of words ambiguity.
	This weakness becomes even more severe when the sentence is wrongly parsed.
\end{itemize}
We hypothesized that the convolution layer helps Constituency Tree-LSTM to mitigate the problem of lacking local context and words ambiguity at leaf nodes.

\section{Combining Convolution and Recursive Neural Networks}\label{sec:cnn-treelstm}

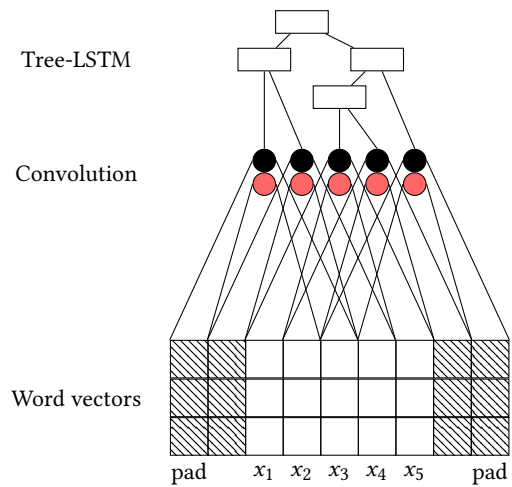
\begin{figure} [H]
    \centering
    \usetikzlibrary{matrix}
\usetikzlibrary{patterns}
\tikzset{
	sq1/.style={rectangle, minimum width=0.5cm, minimum height=0.5cm, text centered, draw=black},
	sq1m/.style={rectangle, minimum width=0.7cm, minimum height=0.3cm, text centered, draw=black},
	sq1p/.style={rectangle, minimum width=0.5cm, minimum height=0.5cm, text centered, draw=black, pattern=north west lines},
	circ/.style={circle, minimum width=0.3cm, minimum height=0.3cm, text centered, draw=black},
	arrow/.style={thick,->},
	sqvec/.style={matrix,matrix of nodes,nodes in empty cells},
}
\tikzstyle{cir} = [circle, minimum width=0.7cm, minimum height=0.7cm, text centered, draw=black ]

\begin{tikzpicture}
\node [sqvec,nodes={circ},      
every even row/.style = { nodes={fill=red!60}},
every odd row/.style = { nodes={fill=black!100}}] (c1) at (1.5,8.5) {
	\\
	\\ 
};  

\node [sqvec,nodes={circ},      
every even row/.style = { nodes={fill=red!60}},
every odd row/.style = { nodes={fill=black!100}}] (c2) at (2,8.5) {
	\\
	\\ 
};  

\node [sqvec,nodes={circ},      
every even row/.style = { nodes={fill=red!60}},
every odd row/.style = { nodes={fill=black!100}}] (c3) at (2.5,8.5) {
	\\
	\\ 
};  

\node [sqvec,nodes={circ},      
every even row/.style = { nodes={fill=red!60}},
every odd row/.style = { nodes={fill=black!100}}] (c4) at (3,8.5) {
	\\
	\\ 
};  

\node [sqvec,nodes={circ},      
every even row/.style = { nodes={fill=red!60}},
every odd row/.style = { nodes={fill=black!100}}] (c5) at (3.5,8.5) {
	\\
	\\ 
};

\node [sqvec,column sep=-\pgflinewidth,nodes={sq1}] (v) at (2.5,5.5) {
	&&&&\\
	&&&&\\
	&&&&\\
};   

\node [sqvec,column sep=-\pgflinewidth,nodes={sq1p}] (v1) at (1,5.5) {
	\\
	\\
	\\
};   
\node [sqvec,column sep=-\pgflinewidth,nodes={sq1p}] (v2) at (0.5,5.5) {
	\\
	\\
	\\
};   
\node [sqvec,column sep=-\pgflinewidth,nodes={sq1p}] (v3) at (4,5.5) {
	\\
	\\
	\\
};   
\node [sqvec,column sep=-\pgflinewidth,nodes={sq1p}] (v4) at (4.5,5.5) {
	\\
	\\
	\\
};   

\draw (v1-1-1.north west) -- (c1-2-1.west); % inner left
\draw (v-1-3.north west) -- (c1-2-1.east); % inner right
\draw (v2-1-1.north west) -- (c1-1-1.west); % outer left
\draw (v-1-4.north west) -- (c1-1-1.east); % outer right

\draw (v-1-1.north west) -- (c2-2-1.west); % inner left
\draw (v-1-4.north west) -- (c2-2-1.east); % inner right
\draw (v1-1-1.north west) -- (c2-1-1.west); % outer left
\draw (v-1-5.north west) -- (c2-1-1.east); % outer right

\draw (v-1-2.north west) -- (c3-2-1.west); % inner left
\draw (v-1-5.north west) -- (c3-2-1.east); % inner right
\draw (v-1-1.north west) -- (c3-1-1.west); % outer left
\draw (v-1-5.north east) -- (c3-1-1.east); % outer right

\draw (v-1-3.north west) -- (c4-2-1.west); % inner left
\draw (v-1-5.north east) -- (c4-2-1.east); % inner right
\draw (v-1-2.north west) -- (c4-1-1.west); % outer left
\draw (v3-1-1.north east) -- (c4-1-1.east); % outer right

\draw (v-1-4.north west) -- (c5-2-1.west); % inner left
\draw (v3-1-1.north east) -- (c5-2-1.east); % inner right
\draw (v-1-3.north west) -- (c5-1-1.west); % outer left
\draw (v4-1-1.north east) -- (c5-1-1.east); % outer right

\node [sq1m] (v8) at (2,10.5) {};
\node [sq1m] (v7) at (1.5,10) {};
\node [sq1m] (v6) at (3,10) {};
\node [sq1m] (v5) at (2.5,9.5) {};
% \draw  (c3) edge (v5);
% \draw  (c4) edge (v5);
\draw  (v5) edge (v6);
% \draw  (c5) edge (v6);
% \draw  (c1) edge (v7);
\draw  (v7) edge (v8);
% \draw  (c2) edge (v7);
\draw  (v6) edge (v8);

\node at (0.5,4.5) {pad};
% \node at (1,4) {pad};
\node at (1.5,4.5) {$x_1$};
\node at (2,4.5) {$x_2$};
\node at (2.5,4.5) {$x_3$};
\node at (3,4.5) {$x_4$};
\node at (3.5,4.5) {$x_5$};
\node at (4.5,4.5) {pad};
% \node at (4.5,4) {pad};

\node at (-1,8.5) {Convolution};
\node at (-1,5.5) {Word vectors};
\node at (-1,10) {Tree-LSTM};

% connect lstm to circle
\draw  (v7) edge (c1-1-1.north);
\draw  (v7) edge (c2-1-1.north);
\draw  (v5) edge (c3-1-1.north);
\draw  (v5) edge (c4-1-1.north);
\draw  (v6) edge (c5-1-1.north);
\end{tikzpicture}
    \caption[CNN-Tree-LSTM]{CNN-Tree-LSTM}
    \label{fig:cnntreelstm} % label MUST come after caption
\end{figure}

Our model architecture is shown in Fig.~\ref{fig:cnntreelstm}.
The model has three modules: word embedding layer, convolution layer, and Constituency Tree-LSTM.

\subsection{Word Embedding Layer}
Suppose that \(Z = \{e_0, e_1, \ldots e_m\}\) is a set input channels with each channel uses a different word embedding.
The first word in a sentence is indexed as word-\(0\)th.
If the sentence is padded with dummy words, left padded dummy words are indexed by negative integers.

For any word embedding~\(e\), the vector presentation of the word-\(i\)th is denoted as~\(w^{(e)}_i \in \mathbb{R}^{d_e}\).
The vector presentation of word-\(i\)th through the set of input channels \(Z\) is expressed as follow:
\begin{align}
 x_i &= w^{(e_0)}_i \ominus w^{(e_1)}_i \ominus  \ldots \ominus w^{(e_m)}_i&\label{concat-emb}
\end{align}
In Eq.\eqref{concat-emb}, \(\ominus\) is concatenation operator which results in the vector \(x_i \in \mathbb{R}^{d}\) with \(d = \sum_{e \in Z} d_e\).
Any sequence of words starting from word-\(i\)th to word-\(j\)th is present as the following matrix:
\begin{align}
X_{i:j} &= x_i \oplus x_{i+1} \oplus \ldots \oplus x_j &\label{concat}
\end{align}
In Eq.\eqref{concat}, \(\oplus\) is concatenation operator which results in the matrix \(X_{i:j} \in \mathbb{R}^{d \times (j-i+1)}\).
\subsection{Convolution Layer}\label{sec:cnn}
Given that \(F\) is the set of all filters of the convolution layer, for any filter \(v \in F\) which has window size \(l\) and set of parameters \(\theta^{(v)} = \{ W^{(v)}, b^{(v)} | W^{(v)} \in \mathbb{R}^{d \times l}, b^{(v)} \in \mathbb{R}\}\), filter \({v}\) is applied on any sequence of word-\(i\)th to word-\((i+l-1)\)th through the following equation:
\begin{align}
c^{(v)}_j &= f(W^{(v)} \otimes X_{i:i+l-1} + b^{(v)}) &\label{filter}
\end{align}
In Eq.\eqref{filter}, operator \(\otimes\) is the Hadamard product~\cite{element-prod}.
\(b \in \mathbb{R}\) is bias term. \(f\) is an activation function.
For indexing, \(j = i + x\) with \(x \in \mathbb{N}\) and \(0 \leq x < l\).
If half-padding policy is employed then \(j = i + \floor{\frac{l}{2}}\).

By slicing the filter \(v\) through the sentence (i.e. applying the filter \(v\) on different sequences of length \(l\) along the sentence) we get vector \(c^{(v)} = [c^{(v)}_0, c^{(v)}_1~\cdots]\) which is a feature map of the sentence~\(s\).
The length of the feature map \(c^{(v)}\) depends on the length of the input sentence, the way which filter \(v\) was slided through the sentence and its window size \(l\)~\cite{conv-arith}.

In our model, all filters in \(F\) are restricted to have odd window sizes and being applied on the sentence according to half padding, unit strides policy~\cite{conv-arith}.
These conditions guarantee that the lengths of all feature maps produced from a sentence are equal to the number of words in that sentence~\cite{conv-arith}.
Suppose the size of the set of filters \(F\) is \(m\), all the feature maps of a sentence of length \(n\) produced by the set of filters is concatenated into one matrix \(P \in \mathbb{R}^{m \times n}\).

The \(i\)-th column vector of \(P\) are treated as the vector representation of the \(i\)-th word in the sentence.

\subsection{Constituency Tree-LSTM}\label{treelstm}
Let \(d\) be the size of the input vectors, \(r
\) be the size of the memory cell and \(z\) be the number of sentiment classes.
\paragraph{Leaf module}
Given any input vector \(x \in \mathbb{R}^d\), the calculation steps inside the leaf module is expressed as follow~\cite{treeLSTM}:
\begin{align}
o &= \sigma{\left( W^{(o)} x + a^{\left(o\right)}\right)} & \\
c &= W^{(c)} x + a^{(c)} & \\
h &= o \odot \tanh{\left(c\right)} &
\end{align}
In this module, \(W^{(o)}, W^{(c)} \in \mathbb{R}^{r \times d}\) and \(a^{\left(o\right)}, a^{(c)} \in \mathbb{R}^r\).
\paragraph{Composer module}
Given the input vectors \({h_l}\) and \({c_l}\) from the left child node, \({h_r}\) and \({c_r}\) from the right child node, the calculation steps inside the composer module are expressed as follow~\cite{treeLSTM}:
\begin{align}
i &= \sigma{ \left(U_l^{(i)} h_{l} + U_r^{(i)} h_{r} + b^{(i)} \right) } &\\
f_{l} &= \sigma{\left(U_{l}^{(l)} h_{l} + U_{r}^{(l)} h_{r} + b^{(f)}\right)} & \\
f_{r} &= \sigma{\left(U_{l}^{(r)} h_{l} + U_{r}^{(r)} h_{r} + b^{(f)}\right)} & \\
o &= \sigma{\left( U_l^{(o)} h_{l} + U_r^{(o)} h_{r} + b^{(o)}\right)} &\\
u &= \tanh{\left( U_l^{(u)} h_{l} + U_r^{(u)} h_{r} + b^{(u)}\right)} &\\
c &= i \odot u + f_{l} \odot c_{l} + f_{r} \odot c_{r} & \\
h &= o \odot \tanh{\left(c\right)} &
\end{align}
For any \(j \in \{i, l, r, o, u\}\) and \(x \in \{l, r\}\), \(U_x^{(j)} \in \mathbb{R}^{r \times r}\) and \( b^{(j)} \in \mathbb{R}^r\).

\paragraph{Output module}
Denoting sequence of words spanned by a sub-tree rooted at node \({j}\) as \({\{x\}_j}\).
Given \({h_j}\) of node \({j}\), the prediction at node \({j}\) is computed by the output module as follow~\cite{treeLSTM}:
\begin{align}
\hat{p_{\theta}}(y \mid \{x\}_j ) &= softmax( W^{(s)} h_j + b^{(s)}) & \\
\hat{y_j} &= \underset{y}{\mathrm{argmax}} \; \hat{p_{\theta}}(y \mid \{x\}_j ) &
\end{align}
With \(W^{(s)} \in \mathbb{R}^{z \times r}\) and \( b^{(s)} \in \mathbb{R}^z\).
\paragraph{Composing sentence}
% Given any sentence, its parse tree and set of vectors representation of each word in the sentence, Constituency Tree-LSTM is applied on the sentence as follow:
% For any non-leaf node, composer module is applied recursively and for any leaf node, the leaf module is applied on the vector representation of the corresponding word.
% Finally, we can apply the output module on any node in the parse tree.
Given any sentence, its parse tree and set of vectors representation of each word in the sentence, Constituency Tree-LSTM is applied on the sentence as follow:
\begin{itemize}
  \item At leaf node, leaf module takes input from previous layer (convolution layer) output of corresponding word.
  \item At non-leaf node, composer module is applied recursively.
  \item After that, at every node \({j}\), output module takes input from leaf module or composer module to predict sentiment of sub-tree root \({j}\).
\end{itemize}

Apart from using Constituency Tree-LSTM to combine the set of vectors representation of each word in a sentence produced be the Convolution layer, we also used LSTM to replace Constituency Tree-LSTM in several experiments.

\subsection{LSTM}\label{sec:lstm}
Denoting input sequence as \(I = \{i_0,\ldots,i_n\}, \forall t, i_t \in \mathbb{R}^n\), LSTM unit~\cite{originLSTM} is expressed as the following recursive formula:
\begin{align}
w_t &= \sigma\left(W^{(w)}i_t + U^{(w)}h_{t-1} + b^{(w)}\right) \label{eq:lstm-input-gate}&\\
f_t &= \sigma\left(W^{(f)}i_t + U^{(f)}h_{t-1} + b^{(f)}\right) \label{eq:lstm-forget-gate}&\\
o_t &= \sigma\left(W^{(o)}i_t + U^{(o)}h_{t-1} + b^{(o)}\right) \label{eq:lstm-output-gate}&\\
u_t &= tanh\left(W^{(u)}i_t + U^{(u)}h_{t-1} + b^{(u)}\right) \label{eq:lstm-update-gate}&\\
c_t &= r_t \odot u_t + f_t \odot c_{t-1} \label{eq:longterm-mem}&\\
h_t &= o_t \odot tanh(c_t) \label{eq:temperal-mem}&
\end{align}
The operation \(\odot\) denotes the element-wise vector product.
Traditionally, \(w_t\), \(f_t\) and \(o_t\) are called input/write gate, forget/deallocate gate and output/read gate respectively and \(c_t\) is called memory cell.
% MAY NOT NECCESSARY
Intuitively, we can interpret how the network works as follow:
\begin{itemize}
  \item \(h_{t-1}\) can be viewed as a short-term memory of the network
  \item \(u_t\) is the information extracted from in current input \(i_t\) and the short-term memory \(h_{t-1}\)
  \item Write gate \(w_t\) decides which information from \(u_t\) will be written into the memory cell \(c_t\)
  \item Forget gate \(f_t\) decides which information will be preserved on memory cell \(c_t\)
  \item \(o_t\) decides which information will be read from the memory cell \(c_t\), which will produce the short-term memory \(h_t\).
\end{itemize}

% in the above formula, \(h_{t-1}\) can be viewed as a short-term memory of the network;
% \(u_t\) is the information extracted from in current input \(i_t\) and the short-term memory \(h_{t-1}\);
% write gate \(w_t\) decides which information from \(u_t\) will be written into the memory cell \(c_t\);
% forget gate \(f_t\) decides which information will be preserved on memory cell \(c_t\);
% and \(o_t\) decides which information will be read from the memory cell \(c_t\), which will produce the short-term memory (or output) \(h_t\).

\section{Transfer Learning From Large Review Dataset}
\subsection{Motivation}\label{sec:we-motive}
There are many words (e.g. ``B-rated'', ``Batman'', ``Nolan'', ``cartoonlike'') which rarely appears in regular documents but more often in movie reviews.
These words might not appear or not have good vector representations in the pre-trained Glove Common Crawl.
Additionally, the ways people use words in movie reviews might be different from their usage in general documents.
For example, the vector presentations of ``sympathy'' and ``disappointed'', or ``boom'' and ``insult'' are very close to each other in Glove Common Crawl but if we have to predict the sentiment of a movie comment which has one of these words.
Ideally, they should be distinctive because ``disappointed'' and ``insult'' are likely to express negative sentiment, while ``boom'' and ``sympathy'' are likely to express positive sentiment.
One solution is to update word embedding during the training process so that we can have better word embedding for specific tasks and domains~\cite{treeLSTM, KimCNN}.
Nevertheless, this method can harm generalization by updating only words which appear in the training set, and thus over-fitting occurs. %and non-related words which only appear in the test set. 
%  
% 
%Noticeably, many experiments~\cite{treeLSTM, KimCNN} have shown that modifying word embeddings during the training process help improving the performance of Deep Learning systems. 
%Although this method improves models' performance, it can harm generalization by updating only words which appear in the training set and non-related words which only appear in the test set. 
% 
 
Moreover, we observed that the available amount of document-level labeled sentiment data (e.g., Amazon Reviews dataset~\cite{amazon-reviews} has 83.68 million reviews) is gigantic compared to the amount of sentence or phrase-level sentiment data (e.g., Stanford Sentiment Treebank~\cite{socher2013recursive} which has 8,544 sentences in its training set, even it is the biggest sentence-level sentiment analysis dataset). 
Our purpose is to utilize this large amount of document-level labeled sentiment data to improve the performance of our models on Stanford Sentiment Treebank. 
 
%
%Noticeably, many experiments~\cite{treeLSTM, KimCNN} have shown that modifying word embeddings during the training process help improving the performance of Deep Learning systems.
%Although this method improves models' performance, it can harm generalization by updating only words which appear in the training set and non-related words which only appear in the test set.

\subsection{Glove Amazon}\label{sec:glove-amazon}
Amazon Reviews~\cite{amazon-reviews} is a gigantic review dataset
which contains 142.8 million reviews from Amazon spanning May 1996 - July 2014\footnote{\url{http://jmcauley.ucsd.edu/data/amazon/}}.
Each review contains product review (rating, text, helpfulness vote) and metadata (descriptions, category information, price, brand, and image features).
The dataset is partitioned into 24 categories (e.g. ``Books'', ``Electronics'', ``Office Products'', ``Movies and TV'').

We hypothesized that by training Glove~\cite{glove} on review documents, especially movie or book reviews, we can capture more rare words and also the different way that people use words (or different word relationships) to express their opinions on movies or books.
This might help our models achieving better generalization when training on small sentence-level sentiment dataset (e.g. Stanford Sentiment Treebank~\cite{socher2013recursive}).
We have five steps to preprocess Amazon dataset for training a new word embedding using Glove method\footnote{Publicly available on Github~\url{https://github.com/stanfordnlp/GloVe}}:
\begin{enumerate}
	\item We only used some partitions of Amazon Reviews dataset which includes:  ``Amazon Movies and TV'' (7,850,072 reviews)~\cite{mcauley2013hidden} and ``Books'' (22,507,155 reviews)~\cite{McAuleyTSH15}~\cite{HeM16}.
	\item All the reviews were grouped by product-ID ( "asin" keyword in the JSON schema of the dataset).
	\item In each product-ID group, the reviews were sorted increasingly by their ratings ("overall" keyword in the JSON schema of the dataset).
	\item All the reviews were dumped into a plain text file.
	\item The text file produced from the previous step was tokenized using Stanford Tokenizer~\cite{tokenizerpart}.
\end{enumerate}
There is no definition of end-of-document in Glove model, which means words which appear in the beginning part of a document will be included in the context of words in the last part of the previous document which leads to noise in training data.
% However, words which appear in the beginning part of a document will be included in the context of words in the last part of the previous document due to no definition of end-of-document in Glove model. As a result, training data might be noisy. Hence, we use step 2 and 3 to mitigate the problem.

We set $x_{max} = 100$, vector size to 300, windows size to 20 and the minimum number of word occurrences to be included in the vocabulary to 5.
The training process took the plain text file from preprocessing steps as input.
In total, the corpus contains 4.7 billion tokens.
After the training process, the resulting word embedding has vocabulary size of 1,734,244.
We named this new word embedding Glove Amazon.

\section{Experiments}
\subsection{Datasets}\label{sec:sst}
We evaluated our model on Standford Sentiment Treebank dataset~\cite{socher2013recursive}.
Standford Sentiment Treebank contains total 11,855 sentences.
We used provided train/dev/test sets which contain 8544, 1101 and 2210 sentences, respectively.
In this dataset, every sentence was parsed using Stanford (constituency) parser~\cite{socher2013recursive} into multiple phrases.
There are a total of 215,154 labeled phrases in the whole dataset. Thus, every sentence in the corpus has a fully labeled parse tree.
For training a Recurrent Neural Network, any phrase spanned by a labeled node is treated as a training sample. 

\paragraph{Fine-grained setting} We partitioned sentiment labels into 5 classes: ``Positive'', ``Somewhat Positive'', ``Neutral'', ``Somewhat Negative'' and ``Negative''.

\paragraph{Binary setting} We removed all ``Neutral'' sentences.
For the remaining 6920/872/1821 sentences in train/dev/test sets, we merged ``Somewhat Positive'' into ``Positive'' and ``Somewhat Negative'' into ``Negative''.

\subsection{Setups}
\subsubsection{Experiment Descriptions}
We did experiment with Glove Amazon and different variations of our model.
\begin{description}
	\item[CNN-Tree-LSTM] Our basic model with only one input channel.
	We initialized word representations with the standard Glove vectors.
	\item[CNN-Tree-LSTM (Glove Amazon)] CNN-Tree-LSTM with word vectors initialized from Glove Amazon.
	\item [2-channel CNN-Tree-LSTM] CNN-Tree-LSTM with two input channels from two different word embedding matrices, which are initialized from Glove Common Craw and Glove Amazon, respectively.
	\item[CNN-LSTM] Similar to CNN-Tree-LSTM. However, we replaced Constituency Tree-LSTM module by an LSTM unit.
	CNN-LSTM has only one input channel initialized from the standard Glove vectors.
	\item [CNN-LSTM (Glove Amazon)] CNN-LSTM with word vectors initialized from Glove Amazon.
	\item [2-channel CNN-LSTM] A CNN-LSTM model with two input channels at CNN layers. The CNN layers are similar to CNN layers in 2-channel CNN-Tree-LSTM model.
	\item[Constituency Tree-LSTM (Glove Amazon)]Glove Amazon~is used to replace the standard Glove vector for initializing~word embedding layer of Constituency Tree-LSTM.
	Apart from that, the whole training process and hyper-parameters of Constituency Tree-LSTM~\cite{treeLSTM} are kept unchanged.
	The case of Tree-LSTM using both Glove Amazon, standard Glove were not experimented on because we have found no reliable way to extend the original Tree-LSTM for multi-channel input.
\end{description}

Since the cost functions of neural networks are nonconvex and algorithms used to train neural networks are only able to find local optimum, different runs of one model can converge in various local optimums depend on the initialized parameters of the models.
Base on the evaluating method used by Tai et al.~\cite{treeLSTM}, we evaluated the above models based on mean, standard deviation of 5 runs.
In addition, the maximum accuracy among 5 runs is also reported.

We index all our experimented models along with their number of parameters in Table.\ref{table:paramtable}.
\begin{table}[H]
	\centering
	\caption{Size of memory cell \(r\) and number of trainable parameters \(\left\vert{\theta}\right\vert\) of our models.
	}
	\label{table:paramtable}
	\begin{tabular}{|l|l|l|}
		\hline
		Model & \(r\) & \multicolumn{1}{|c|}{\(\left\vert{\theta}\right\vert\)}\\ \hline
		CNN-LSTM                 & 168         & 489,347          \\
		CNN-Tree-LSTM            & 150         & 482,153          \\
		2-channel CNN-LSTM       & 168         & 729,347          \\
		2-channel CNN-Tree-LSTM  & 150         &
		722,153 \\
		\hline
	\end{tabular}
\end{table}
\subsubsection{Hyper-parameters and Training}
We trained models on training set and tuned hyper-parameters on development set of Stanford .
Our models was trained using AdaGrad~\cite{duchi2011adaptive} with learning rate of $\{0.1, 0.05, 0.01\}$, L2 regularization strength of $\{1e^{-3},~ 1e^{-4}, ~ 1e^{-5} \}$ and batch size of 25.
Word vectors are updated with learning rate $\alpha$ of $\{0.1,~0.05, ~0.01\}$.
In convolution layers, treated number of filter and filter size as hyper-parameters.
We found that 100 filters of size 3 and 100 filters of size 5 yield better results compared to single filters size or the number of filters larger than 200.
We regularized the convolution layers with input dropout rate of 0.5, and output dropout rate of 0.2. At output layer, we regularized with dropout rate of 0.5.
Training with Adagrad's learning rate of 0.01 and word vectors' learning rate 0.1 give the best result.

Our models were trained for 60 epochs.
\subsection{Results}\label{sec:result}
\begin{table*}[]
	\centering
	\caption[Experiment result on SST]{
		Experiment results of models evaluated on Stanford Sentiment Treebank.
		The accuracies of models in blocks A to E are taken from their original papers.
		We highlight the best results among our models and underline the state-of-the-art results.
		\textit{(*): Mean and standard deviation of 5 runs.}
		\textit{(**): Mean of 100 runs, std was not reported.}
	}
	\label{table:experimentresult}
	\begin{tabular}{|c|l|ll|ll|}
		\hline
		\textbf{Block} & \textbf{Model}  & \multicolumn{2}{c|}{\textbf{Binary}} & \multicolumn{2}{c|}{\textbf{Fine-grained}}  \\
		\Xhline{3\arrayrulewidth}
		\Xhline{3\arrayrulewidth}

		\multirow{4}{*}{A} & CNN-non-static~\cite{KimCNN} & \multicolumn{2}{c|}{87.2} & \multicolumn{2}{c|}{48.0} \Tstrut \\
		& CNN-multichannel~\cite{KimCNN} & \multicolumn{2}{c|}{88.1} & \multicolumn{2}{c|}{47.4} \\
		& DCNN~\cite{DCNN} & \multicolumn{2}{c|}{86.8} & \multicolumn{2}{c|}{48.5} \\
		& MVCNN~\cite{2-layer-cnn} & \multicolumn{2}{c|}{89.4} & \multicolumn{2}{c|}{49.6} \\
		\hline
		\multirow{6}{*}{B} & LSTM~\cite{treeLSTM}   & \multicolumn{2}{c|}{84.9 (0.6)*} & \multicolumn{2}{c|}{46.4 (1.1)*} \\
		& BiLSTM~\cite{treeLSTM}  & \multicolumn{2}{c|}{87.5 (0.5)*} & \multicolumn{2}{c|}{49.1 (1.0)*}   \\
		& 2-layer LSTM~\cite{treeLSTM} & \multicolumn{2}{c|}{86.3 (0.6)*} & \multicolumn{2}{c|}{46.0 (1.3)*} \\
		& 2-layer Bidirectional LSTM~\cite{treeLSTM} & \multicolumn{2}{c|}{87.2 (1.0)*} & \multicolumn{2}{c|}{48.5 (1.0)*} \\
		& DMN~\cite{attention-gru} & \multicolumn{2}{c|}{88.6 } & \multicolumn{2}{c|}{52.1} \\
		& Byte mLSTM~\cite{mlstm} & \multicolumn{2}{c|}{\underline{91.80}**} & \multicolumn{2}{c|}{52.90**} \\
		\hline
		\multirow{6}{*}{C} & RNTN~\cite{socher2013recursive}  & \multicolumn{2}{c|}{85.4} & \multicolumn{2}{c|}{45.7} \\
		& DRNN~\cite{IrsoyDRNN} & \multicolumn{2}{c|}{86.6} & \multicolumn{2}{c|}{49.8}  \\
		& TE-RNTN~\cite{tag-embedding-rnn} & \multicolumn{2}{c|}{87.7} & \multicolumn{2}{c|}{48.9} \\
		& Dependency Tree-LSTM~\cite{treeLSTM}  & \multicolumn{2}{c|}{85.7 (0.4)*} & \multicolumn{2}{c|}{48.4 (0.4)*}  \\
		& Constituency Tree-LSTM~\cite{treeLSTM} &  \multicolumn{2}{c|}{88.0 (0.3)*}  & \multicolumn{2}{c|}{51.0 (0.5)*} \\
		& Constituency Tree-LSTM Ensemble~\cite{LooksHHN17} & \multicolumn{2}{c|}{90.2} & \multicolumn{2}{c|}{\underline{53.6}} \\
		\hline
		\multirow{3}{*}{D} & GICF~\cite{group-instance} & \multicolumn{2}{c|}{85.7}  &  \multicolumn{2}{c|}{-} \\
		& Paragraph-Vec~\cite{ParagraphVec} & \multicolumn{2}{c|}{87.8} & \multicolumn{2}{c|}{48.7} \\
		& LSTM (PARAGRAM-SL999)~\cite{wieting2015towards} & \multicolumn{2}{c|}{89.2} & \multicolumn{2}{c|}{-}
		\\
		\hline
		\multirow{2}{*}{E}  & CNN-GRU-word2vec~\cite{cnn-rnn}                    &\multicolumn{2}{c|}{89.95} & \multicolumn{2}{c|}{50.68} \\
		& CNN-LSTM-word2vec~\cite{cnn-rnn}   &       \multicolumn{2}{c|}{89.56} & \multicolumn{2}{c|}{51.50} \Bstrut    \\
		\Xhline{3\arrayrulewidth}
		\Xhline{3\arrayrulewidth}
		&   & \textbf{Mean(std)} & \textbf{Max} & \textbf{Mean(std)} & \textbf{Max}  \\
		\cline{3-6}
		\multirow{6}{*}{F} & Constituency Tree-LSTM ~\cite{treeLSTM} (Glove Amazon) & 88.85 (0.44) & 89.35 & 50.53 (0.98) & 51.31 \Tstrut \\
		& CNN-LSTM                                 & 89.10 (0.39)  & 89.40 & 51.92 (0.63) & 52.66 \\
		& CNN-LSTM (Glove Amazon) & 89.25 (0.73) & \textbf{90.39}  & 50.84 (0.79) & 51.85 \\
		& 2-channel CNN-LSTM                        & 89.44    (0.51) & 90.01 & 51.70 (0.57) & 52.53 \\
		& CNN-Tree-LSTM                            & 88.82 (0.13) & 88.92 & 51.35 (1.45) & 52.94 \\
		& CNN-Tree-LSTM (Glove Amazon)             & 88.96 (0.24) & 89.18 & 51.51 (0.99) & 52.80 \\
		& 2-channel CNN-Tree-LSTM  & \textbf{89.70 (0.36)} & 90.12  & \textbf{52.46 (0.55)} & \textbf{53.03} \Bstrut  \\
		\hline
	\end{tabular}
\end{table*}
Experiment results are summaries in Table \ref{table:experimentresult}.
Table \ref{table:experimentresult} contains two parts.
Block A to E contain all baselines model.
Block F contains all models we proposed and evaluated.
\begin{description}
	\item[Block A] contains convolution neural networks.
	CNN-non-static and CNN-multichannel~\cite{KimCNN} are single layer CNN.
	DCNN~\cite{DCNN} and MVCNN~\cite{2-layer-cnn} are multilayer CNNs, with MVCNN is a large model which has 2 layers and 5 input channels.
	\item[Block B] contains recurrent neural network models and their variations.
	All the models in this Block B process sentences sequentially.
	DMN~\cite{attention-gru} is a sophisticated model used GRU with attention mechanism and episodic memory.
	Byte mLSTM~\cite{mlstm} is the state-of-the-art system on binary setting.
	\item[Block C] contains models which belong to the family of Recursive Neural Networks (tree-structured models).
	RNTN~\cite{socher2013recursive} is the first recursive neural network to successfully apply on sentence-level sentiment analysis (Stanford Sentiment Treebank).
	DRNN~\cite{IrsoyDRNN} is a multilayered extension of RNTN.
	TE-RNTN is also an extension of RNTN which utilize the local syntactic information at each node of a sentence's parse tree.
	Constituency Tree-LSTM Ensemble~\cite{LooksHHN17} is an ensemble of 30 Constituency Tree-LSTMs.
	This model is the state-of-the-art system on the fine-grained setting of Stanford Sentiment Treebank.
	\item[Block D] contains transfer learning methods, which utilized a large amount of data other than Stanford Sentiment Treebank.
	GICF~\cite{group-instance} learns to classify sentiments of sentences (in Stanford Sentiment Treebank) using only document-level sentiment labels training dataset.
	Paragraph-Vec~\cite{ParagraphVec} learns to encode any sequence of words into a vector with the purpose of maximizing the likelihood of words which appear in that sequence given the encoding vector.
	\item[Block E] contains models which combine Convolution Neural Networks and Recurrent Neural Networks.
\end{description}
\subsection{Discussion}
\subsubsection{Combination of Convolution and Recursive Neural Networks}
The fact that CNN-Tree-LSTM outperforms Constituency Tree-LSTM~\cite{treeLSTM} and CNN-multichannel~\cite{KimCNN} supports our hypothesis on the benefits of combining convolution layers with Tree-LSTM.
Additionally, the combination of CNN and LSTM or TreeLSTM outperforms most Convolution Network Networks in Block A. Furthermore, the results support our hypothesis that max pooling layer can be harmful to CNN by ignoring the position of features.

\subsubsection{Glove Amazon versus standard Glove}
\begin{figure} []
	\centering
	\pgfplotstableread[row sep=\\,col sep=&]{
    interval & CNN LSTM & CNN TreeLSTM & TreeLSTM \\
    StandardGlove  & 89.1  &  88.82  & 88\\
    GloveAmazon   & 89.25 & 88.96 & 88.85\\
    Combination  & 89.44  & 89.70 &   \\
    }\mydata

\begin{tikzpicture}[scale=1]
    \begin{axis}[
            ybar,
			bar width=0.6cm,
			width=0.5\textwidth,
			height=0.37\textwidth,
			legend style={at={(0.5,-0.1)},
				anchor=north,legend columns=-1},
            symbolic x coords={StandardGlove, GloveAmazon, Combination},
            xtick=data,
            nodes near coords,
            enlarge x limits={0.2},
            nodes near coords align={vertical},
            ymin=87.8,ymax=90,
            ylabel={},
        ]
        \addplot table[x=interval,y=CNN LSTM]{\mydata};
        \addplot table[x=interval,y=CNN TreeLSTM]{\mydata};
        \addplot table[x=interval,y=TreeLSTM]{\mydata};
        \legend{CNN LSTM, CNN TreeLSTM, TreeLSTM}
    \end{axis}
\end{tikzpicture}
	\caption[qwerty]{Mean accuracy of 5 runs on binary setting of Tree-LSTM, CNN-LSTM and CNN-Tree-LSTM using different word embeddings}
	\label{graph:binary}
\end{figure}
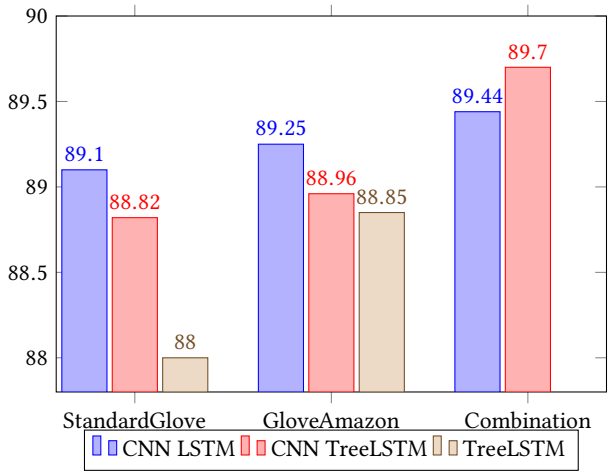
On binary setting (Fig. ~\ref{graph:binary}), Glove Amazon helps single input-channel models to achieve higher accuracy compared to the standard Glove.
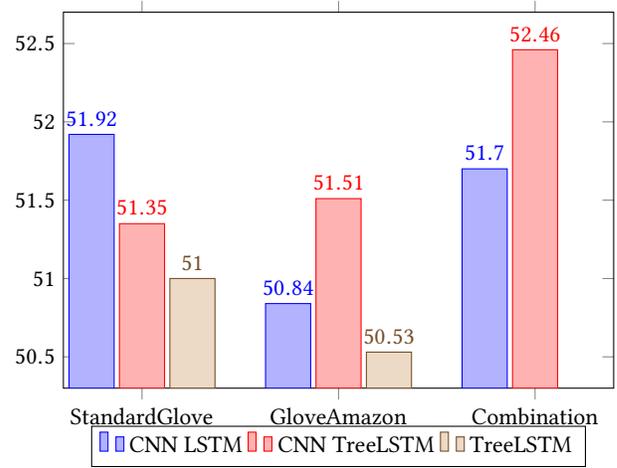
\begin{figure} []
	\centering
	\pgfplotstableread[row sep=\\,col sep=&]{
    interval & CNN LSTM & CNN TreeLSTM & TreeLSTM \\
    StandardGlove  & 51.92  &  51.35  & 51\\
    GloveAmazon   & 50.84 & 51.51 & 50.53\\
    Combination  & 51.7  & 52.46 &   \\
    }\mydata

\begin{tikzpicture}[scale=1]
    \begin{axis}[
            ybar,
			bar width=0.6cm,
			width=0.5\textwidth,
			height=0.37\textwidth,
			legend style={at={(0.5,-0.1)},
				anchor=north,legend columns=-1},
            symbolic x coords={StandardGlove, GloveAmazon, Combination},
            xtick=data,
            nodes near coords,
            enlarge x limits={0.2},
            nodes near coords align={vertical},
            ymin=50.3,ymax=52.7,
            ylabel={},
        ]
        \addplot table[x=interval,y=CNN LSTM]{\mydata};
        \addplot table[x=interval,y=CNN TreeLSTM]{\mydata};
        \addplot table[x=interval,y=TreeLSTM]{\mydata};
        \legend{CNN LSTM, CNN TreeLSTM, TreeLSTM}
    \end{axis}
\end{tikzpicture}
	\caption[qwerty]{Mean accuracy of 5 runs on fine-grained setting of Tree-LSTM, CNN-LSTM and CNN-Tree-LSTM using different word embeddings}
	\label{graph:fine-grained}
\end{figure}
However, on fine-grained setting (Fig. ~\ref{graph:fine-grained}), Glove Amazon is worse than the standard Glove in most cases except for CNN-Tree-LSTM.
In short, Glove Amazon is good for models on binary setting however harmful in the fine-grained setting.
%The reason is that our Glove Amazon is trained on review dataset, words which are used to express the same sentiment are more likely to co-occur.
The reason is that in review dataset that we trained Glove Amazon on, words which are used to express the same sentiment are more likely to co-occur.
Therefore, the Glove Amazon word vectors are more dependent on the sentiment expressed in the words compared to the word vectors in the standard Glove which capture more general meaning of words.
% This possibly is the reason why Glove Amazon is beneficial for classifying positive and negative sentiments in binary setting but harmful to detecting neutral in the fine-grained setting.

A combination of Glove Amazon and the standard Glove improves the accuracy of both 2-channel CNN-LSTM and 2-channel CNN-Tree-LSTM on fine-grained setting as well as 2-channel CNN-Tree-LSTM on the binary setting.
Yet, the improvements of CNN-Tree-LSTM are more significant compared to CNN-LSTM.
\subsubsection{Tree-structured Versus Sequential Models}
With single input channel, CNN-LSTM outperforms CNN-Tree-LSTM in most cases except the case of using Glove Amazon on the fine-grained setting.
With two input channels using both Glove Amazon and the standard Glove, CNN-Tree-LSTM gains a large improvement and outperforms 2-channel CNN-LSTM on both binary and fine-grained setting.
On average, 2-channel CNN-Tree-LSTM achieves the highest accuracy among all our models in both settings.

\subsubsection{Comparing to the State-of-the-art Models}
Constituency Tree-LSTM Ensemble~\cite{LooksHHN17} is the state-of-the-art model on the fine-grained setting.
Although the performance of 2-channel CNN-Tree-LSTM is lower then Constituency Tree-LSTM Ensemble, Constituency Tree-LSTM Ensemble is not an novel model and is an ensemble of \(30\) Constituency Tree-LSTMs.
Since, the main purpose of Looks et al. were to use Constituency Tree-LSTM as an example to demonstrate the concise and batch-wise parallelism of its TensorFlow Fold implementation~\cite{LooksHHN17}.

2-channel CNN-Tree-LSTM is comparable with Byte mLSTM~\cite{mlstm} on the fine-grained setting but underperforms Byte mLSTM in the binary setting.
Similar to our approach, Radford st al. did transfer learning from large review dataset (which is also Amazon Reviews dataset) to improve the performance of their models on Stanford Sentiment Treebank.
But different from our work, they trained a byte-level multiplicative LSTM~\cite{KrauseLMR16} language model on Amazon Reviews dataset.
After that, the vector representations of all sentences in Stanford Sentiment Treebank were computed using the trained byte-level multiplicative LSTM.
These vector representations along with their corresponding labels were used to train a logistic regression classifier~\cite{mlstm}.

On Stanford Sentiment Treebank, their method achieved state-of-the-art performance on the binary setting and comparable performance with Constituency Tree-LSTM Ensemble on fine-grained setting.
Despite that, the performances of their model on other NLP tasks were not impressive, these tasks include: semantic relatedness (SICK~\cite{SemeEvalTask1}), subjectivity/objectivity detection (SUBJ~\cite{cs-CL-0409058}), opinion polarity (MPQA~\cite{Wiebe2005}) and paraphrase detection (Microsoft Paraphrase Corpus~\cite{Dolan:2004:UCL:1220355.1220406}).
It is likely that these tasks are out-of-domain for their model which only trained on review dataset~\cite{mlstm}.

While conducting this research, we have trained CNN-LSTM as language model on Amazon Reviews dataset.
Nevertheless, we have not gained any success with this method.

\section{Conclusion}
In this paper, we introduced a combination of Recursive Neural Network and Convolution Neural Network for sentence-level sentiment analysis.
We experimented with both tree-structured and sequential Recursive Neural Networks.
Using Standford Sentiment Treebank, we demonstrated that Recurrent or Recursive Neural Networks can be used for combining phrase-level features produced by Convolution Neural Networks.
Our experiments show that these combinations outperform most pure Convolution, Recurrent and Recursive Neural Networks.
These results provide further support for the hypothesis that the usage of Recurrent or Recursive Neural Networks is better than k-max-pooling layer in the respect of preserving features' position information and capturing long-range dependencies between features.

Additionally, in an attempt to improve vector presentations of words, we trained Glove vectors on the gigantic Amazon Reviews dataset (Glove Amazon).
We have demonstrated Glove Amazon is good for these models on the binary setting but can be harmful to them in the fine-grained setting.
Our experiments also show that a combination of both Glove Amazon and the standard Glove is more beneficial for CNN-Tree-LSTM compared to CNN-LSTM as 2-channel CNN-Tree-LSTM outperforms 2-channel CNN-LSTM on both binary and fine-grained setting.

\appendix
%Appendix A
%
\begin{acks}
We gratefully acknowledges the support of VNG Corporation and research funding from Advanced Program in
Computer Science, University of Science, Vietnam National University - Ho Chi
Minh City. 
\end{acks}

\bibliographystyle{ACM-Reference-Format}
\bibliography{sample-bibliography}

\end{document}